# NeuroWrite: "Predictive Handwritten Digit Classification using Deep Neural Networks."

KOTTAKOTA ASISH
*Computer Science and Engineering*
*SRM Institute of Science and Technology*
*Ramapuram, Chennai, India*
ka9620@srmist.edu.in

P. SARATH TEJA
*Computer Science and Engineering*
*SRM Institute of Science and Technology*
*Ramapuram, Chennai, India*
ps8227@srmist.edu.in

R. KISHAN CHANDER
*Computer Science and Engineering*
*SRM Institute of Science and Technology*
*Ramapuram, Chennai, India*
kr3265@srmist.edu.in

Dr.D. Deva Hema
*Computer Science and Engineering*
*SRM Institute of Science and Technology*
*Ramapuram, Chennai, India*
devahemd@srmist.edu.in

**Abstract**— The rapid evolution of deep neural networks has revolutionized the field of machine learning, enabling remarkable advancements in various domains. In this article, we introduce NeuroWrite, a unique method for predicting the categorization of handwritten digits using deep neural networks. Our model exhibits outstanding accuracy in identifying and categorising handwritten digits by utilising the strength of convolutional neural networks (CNNs) and recurrent neural networks (RNNs).In this article, we give a thorough examination of the data preparation methods, network design, and training methods used in NeuroWrite. By implementing state-of-the-art techniques, we showcase how NeuroWrite can achieve high classification accuracy and robust generalization on handwritten digit datasets, such as MNIST. Furthermore, we explore the model's potential for real-world applications, including digit recognition in digitized documents, signature verification, and automated postal code recognition. NeuroWrite is a useful tool for computer vision and pattern recognition because of its performance and adaptability.The architecture, training procedure, and evaluation metrics of NeuroWrite are covered in detail in this study, illustrating how it can improve a number of applications that call for handwritten digit classification. The outcomes show that NeuroWrite is a promising method for raising the bar for deep neural network-based handwritten digit recognition.

## I. INTRODUCTION

The field of handwritten digit recognition is complex and varied, requiring careful execution. It includes the use of tried-and-true algorithms, dealing with huge datasets, and putting feature extraction and scaling methods to use. At its core, this pursuit revolves around employing Neural Networks to discern handwritten digits within scanned images, empowering machines to decipher human-generated numerals. Nevertheless, this research domain is inherently challenging due to the inherent idiosyncrasies of human handwriting. The uniqueness of each individual's handwriting presents a formidable obstacle, requiring explicit training for computers to recognize digits effectively. Consequently, the pivotal role of deep learning becomes evident in addressing issues spanning image classification, sound recognition, object detection, image segmentation, and object recognition, particularly when distinguishing visually similar digits such as 1 and 7, 5 and 6, 3 and 8, among others. The added complexity introduced by variations in handwriting among different individuals further exacerbates the challenge, as the distinctiveness of each writing style plays a significant role in numeral formation. Consequently,this paper introduces deep learning techniques and principles as a means to confront and surmount these formidable challenges. This predictive handwritten digit classification system harnesses the power of Neural Networks, involving the loading of images from .png files while extracting features and labels. These features are appropriately rescaled by dividing them by 255 to prevent computational overflow. The dataset comprises 10,000 testing cases and 60,000 training examples. To minimize errors, a hypothesis is systematically applied to the network's layers.

At the heart of solving this challenge lies the application of Neural Networks, which leverage a wealth of handwritten numeral instances or training examples. These instances enable the network to learn and autonomously identify handwritten digits. Furthermore, increasing the quantity of training examples enriches the network's understanding of handwriting styles, consequently enhancing accuracy.

## II. LITERATURE REVIEW

Sandeep Dwarkanath Pandea et al. [1] have introduced a novel approach known as Devanagari Handwritten Text Recognition (DHTR). They utilize a Convolutional Neural Network (CNN) to enhance recognition rates and automate the digitization of the Devanagari script, offering a solution for preserving ancient medical prescriptions and treatments from Vedic literature. Their work, which focuses on optimizing both recognition accuracy and conflict resolution, demonstrates the potential to transform handwritten Devanagari text into digital format, significantly reducing the storage space and retrieval challenges associated with paper-based documents.

Sandeep Dwarkanath Pandea and the team effectively leverage the Devanagari Handwritten Character Dataset (DHCD), comprising numerous Devanagari script characters. By explicitly addressing the need for digitization tools for Devanagari, their research opens doors to improved accessibility and manipulation of valuable data, making it a crucial step in the transition from manual to automated systems for Devanagari script in India.

Mohamed Ali Souibgui et al. [2] present an efficient approach to handwriting recognition using few-shot learning. By using a minimal number of photos for each alphabet symbol, this technique reduces the need for substantial human annotation.
It provides a workable way for identifying alphabets in textline photographs, especially in low-resource situations with uncommon scripts. The model automatically assigns pseudo-labels to unlabeled data through the use of unsupervised progressive learning, minimising the labor-intensive process of manual labelling. This strategy maintains excellent accuracy while easing the load of manual labelling, which is especially useful for Handwritten Text Recognition (HTR) when working with sparse data or old scripts.

In situations where labeled data is limited, training data-intensive deep learning models for handwritten text recognition can be challenging, particularly with obscure or encrypted scripts. Historical ciphers with complex alphabets further complicate the task. Unsupervised methods, while useful with scarce labeled data, often yield lower performance, while supervised approaches excel but require significant amounts of labeled data. Few-shot learning, as exemplified by Mohamed Ali Souibgui's character matching strategy, offers a promising alternative, delivering high accuracy while alleviating the manual labeling workload.

Ali Alameer et al. [3] introduced a novel approach using dictionary learning for handwritten number recognition. They harnessed the power of class-specific dictionaries, adopting the histogram of oriented gradients (HOG) feature space for fine-grained details in the images. Their approach demonstrated enhanced classification performance across diverse languages and highlighted the effectiveness of combining HOG features with dictionary learning while using fewer parameters than existing deep learning models.

In various applications, pattern recognition, including person identification and image classification, is crucial. Recent methods have integrated neural networks, such as LeNet-5 and support vector machines (SVM), to classify handwritten digits. Some promising results have emerged from combining HOG and SVM for feature extraction in different languages, although HOG's effectiveness in recognizing handwritten Chinese characters deserves further exploration.

A new approach for Handwritten Number Recognition (HNR) is presented by M. I. R. Shuvo et al. [4] and is based on the idea that handwritten numerals are distinct deformations of printed forms. When handwritten numeral images are superimposed on their corresponding printed numeral images, this hypothesis makes recognition tasks easier and increases accuracy. Their suggested technique converts handwritten number images (HNIs) into printed number images (PNIs) using auto-encoders and convolutional auto-encoders. Neural networks and convolutional neural networks are modified for classification. For classifying numeral categories, Support Vector Machine (SVM) and other classification algorithms are used. The HNR system achieves remarkable recognition accuracy for Bengali, Devanagari, and English handwritten numerals, reaching 99.68%, 99.73%, and 99.62%, respectively. The superimposition technique reduces computational overhead.

This study introduces a successful HNR system that was motivated by the idea of people learning to write. For quick identification, printed forms can have handwritten numbers, which are distinctive interpretations of printed numbers, superimposed on them. By creating a superimposition module, this method streamlines recognition from Printed Number Images (PNIs) and serves as a straightforward image classification task, aiming to enhance handwritten number recognition by transforming the way printed and handwritten forms interact.

In their work, Hongge Yao et al. [5] develop FOD_DCNet, a deep capsule network designed for recognizing and separating fully-overlapping handwritten digits. FOD_DCNet employs small convolution kernels to extract fine-grained features efficiently while reducing the number of parameters. They introduce the concept of capsule dimensions to better represent extracted features, minimizing information loss. Additionally, they propose a "series dual dynamic routing collocation" to optimize routing for classification. Compared to CapsNet, FOD_DCNet significantly improves classification efficiency, achieving a remarkable 93.53% accuracy – 5.43% higher than CapsNet, with only 55.61% of the parameters.

The recognition and separation of fully-overlapping handwritten digits pose a substantial challenge for computer vision systems. Human eyes can easily identify overlapping characters, but this task is complex for computers due to the need for precise understanding of stroke details and digit positioning. Despite the advancements in convolutional neural networks, effectively addressing this issue remains a significant benchmark for network performance and provides a foundation for tackling similar challenges in other languages, such as English and Chinese. FOD_DCNet's remarkable detection and separation accuracy highlight its effectiveness in handling overlapping digits, aided by the robust positional links between capsule vectors, enabling the extraction and comprehension of diverse features.

Fatemeh Haghighi et al. [6] present a novel model for recognizing handwritten digits, employing a stacking ensemble classifier. This innovative approach combines convolutional neural networks (CNN) with bidirectional long-short term memory (BLSTM) and utilizes the probability vector of image class as input for meta-classifier layers. Leveraging the BLSTM's capability to understand arrays and vectors, this model enhances deep learning accuracy by incorporating the output probability vector from the first model. To ensure reliable results, the model is extensively tested on a large Persian/Arabic dataset, accommodating various writing styles from different individuals, with remarkable accuracy rates of 99.98% in training and 99.39% in testing. The approach's emphasis on structural similarities within Persian/Arabic numerals, along with its handling of variations in writing styles, contributes to its overall success.

Amirreza Fateh et al. [7] tackle the challenge of multilingual handwritten numeral recognition with a language-independent model based on a robust CNN. This model encompasses both language recognition and digit recognition, serving critical roles in applications like address sorting and license plate recognition. Given the international nature of modern communication and transactions in multilingual regions, systems capable of recognizing multiple languages are invaluable. Handwritten numeral recognition is inherently more complex than recognizing printed numerals due to diverse and intricate handwriting styles. The proposed system employs transfer learning to enhance image quality and recognition performance. Extensive experiments across six different languages yield an average accuracy of up to 99.8%, demonstrating the model's robustness and cost-effective applicability for recognizing handwritten numerals in various languages.

Abhinandan Chiney et al. [8] have developed an efficient system for handwritten English character recognition in manually filled forms. They created a comprehensive dataset comprising 84,712-character images by collecting and labeling handwritten characters from 572 forms filled by over 200 individuals, introducing demographic diversity. This dataset, known as HW-dataset, includes both alphabetical and numerical characters. Three hybrid datasets (h-EHd, h-EHa, h-EHm) were formed by combining the HW-dataset with the EMNIST dataset.

The authors utilized an anchor-based image extraction technique and trained a Multi-Channel CNN model on the hybrid datasets to automate the digitization of handwritten forms. The classification accuracy of the MCCNN for h-EHa, h-EHd, and h-EHm is notably high, at 93%, 96%, and 93% on test data, outperforming models trained solely on the EMNIST dataset. This innovative model proves highly effective in digitizing handwritten forms and is particularly suitable for touch-free document handling.

Savita Ahlawat and her team [9] developed a hybrid model, combining Convolutional Neural Networks (CNN) and Support Vector Machine (SVM), to recognize handwritten digits from the MNIST dataset. In this hybrid approach, CNN automatically extracted features from diverse and distorted handwritten digits, while SVM served as a binary classifier. The experimental results demonstrated the model's effectiveness, achieving an impressive 99.28% recognition accuracy on the MNIST dataset. Unlike manual feature extraction, this non-handcrafted approach directly extracts features from original images, eliminating the need for prior human knowledge and saving time. By leveraging the CNN model, topological information is considered, making it robust to transformations like rotation and translation. Section 2 provides an overview of related studies in this field, while Section 3 details the proposed methodology. The remaining sections cover experimental design, results, conclusion, and future directions for this handwritten digit recognition system.

Anisha Gupta et al. [10] introduced a novel multi-objective optimization framework to efficiently identify informative local regions in character images. The approach considered three key objectives: 1) recognition accuracy, 2) average recognition time per character image, and 3) redundancy of local regions. Identifying these regions is crucial for robust handwritten character recognition. The authors employed a modified opposition-based multi-objective Harmony Search algorithm to rank local regions within a 3D Pareto front. Their method was tested on four datasets, including isolated Bangla Basic characters, Bangla numerals, English numerals, and isolated Devanagari characters. The results demonstrated a significant reduction in recognition costs and redundancy while improving recognition accuracy, offering a cost-effective solution for isolated handwritten character recognition.

In handwritten character recognition, a common approach involves dividing character images into localized zones. Each local zone's attributes contribute to the final classification. Two main methods for zone division are dynamic and static zoning. Static zoning generates fixed-size local zones with predefined shapes, sizes, and positions, while dynamic zoning relies on statistical or topological aspects of the character image. This summary highlights both static and dynamic zoning techniques. Region-based sampling efficiently identifies distinct spatial regions within a collection of images, offering cost savings at the expense of increased time.

## III. METHODOLOGY

A dataset, such as the MNIST dataset, comprising 70,000 grayscale images of handwritten digits, serves as the foundation for image processing, an ever-expanding field that involves importing, analyzing, and altering images to extract valuable information or enhance their quality. In this process, data undergoes three key phases: pre-processing, augmentation, and information extraction. Neural networks, an integral component of deep learning, emulate the human brain's structure and communication, using interconnected layers of artificial neurons to process data. These networks require training data to refine their accuracy, making them powerful tools for tasks like speech and image recognition. The MNIST dataset plays a crucial role in training and testing such neural networks for handwritten digit recognition, enabling the development of advanced classification models

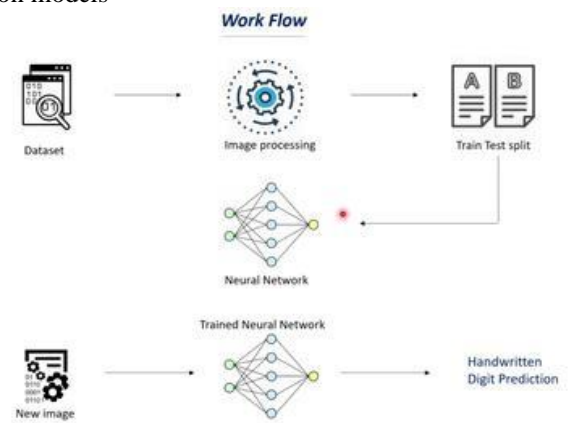

### Dataset

In the fields of machine learning and computer vision, the MNIST dataset is a frequently used dataset. For "Modified National Institute of Standards and Technology" dataset. The MNIST dataset of handwritten digits is frequently used for testing and refining different machine learning and deep learning algorithms. It was made by altering a portion of the NIST dataset (National Institute of Standards and Technology). The MNIST dataset consists of the following key characteristics:

**Images:** It contains a set of 28x28 pixel grayscale images, each representing a single digit (0 through 9).

**Labeling:** Each image is associated with a label, which indicates the digit that the handwritten image represents. This label is an integer between 0 and 9

**Size:** The dataset typically has 10,000 test images and 60,000 training images.

## Data Preparation

**Data Loading:** The MNIST dataset or other handwritten digit datasets are loaded, and the images are preprocessed. Preprocessing may involve resizing, grayscale conversion, and other image enhancements.

**Data Splitting:** Typically, training, validation, and testing sets are separated from the dataset. The neural network is trained using the training set, validated using the validation set, and tested against the testing set to determine how well the model performed.

## Neural Network Architecture

**Input Layer:** The input layer has neurons corresponding to the number of features in an image (e.g., pixels in the case of MNIST).

**Hidden Layers**: One or more hidden layers contain neurons or nodes that learn patterns and features from the input data. The architecture and depth of these layers can vary depending on the specific neural network design.

**Output Layer:** The output layer typically contains 10 neurons (for digits 0-9), each representing the likelihood that the input digit belongs to a specific class.

**Activation Functions:** Non-linear activation functions (e.g., ReLU, Sigmoid, or Tanh) are applied to the neurons in the hidden layers to introduce non-linearity into the model, allowing it to capture complex patterns in the data.

**Training**: The neural network is trained using the training data, which involves feeding the input data through the network, calculating the loss, and backpropagating the gradients to update the model's parameters. This process is repeated for multiple epochs until the model converges.

**Validation:** To avoid overfitting, the model's performance is tracked during training on the validation set. Based on the validation results, hyperparameters can be changed, including learning rate, number of layers, and number of neurons in each layer.

**Testing and Evaluation:** The model is evaluated on the testing dataset after training to determine its accuracy and ability to generalize to new data.

## IV. WORKING

### 1. Loading the MNIST Dataset:

The code starts by loading the MNIST dataset using TensorFlow and Keras. This dataset consists of a large collection of images of handwritten digits (0 to 9) utilized for training and testing the model.

```
# Load the MNIST dataset
mnist = keras.datasets.mnist
(train_images, train_labels), (test_images, test_labels) = mnist.load_data()
```
✓ 0.6s

```
print(train_images.shape ,train_labels.shape,test_images.shape,test_labels.shape)
```
✓ 0.0s

(60000, 28, 28) (60000,) (10000, 28, 28) (10000,)

- Train Data = 60,000 images
- Test Data = 10,000 images
- Image Dimension - 28*28
- GrayScale Image - 1 Channel

Fig.1 Load Data

### 2. Training the Neural Network:

Convolutional layers for feature extraction, max-pooling layers for downsampling, fully connected layers for classification, and dropout layers to avoid overfitting are all included in the model architecture. The model's weights are modified while it is being trained using the 'adam' optimizer. For multi-class classification tasks with integer labels, the 'sparse_categorical_crossentropy' loss function is employed. The model is trained for 10 epochs, with a batch size of 64, and a 20% validation split to monitor the training process.

```python
# Preprocess the data
train_images = train_images / 255.0
test_images = test_images / 255.0
```
✓ 0.2s

```python
# Build a CNN model
model = keras.Sequential([
    keras.layers.Conv2D(32, (3, 3), activation='relu', input_shape=(28, 28, 1)),
    keras.layers.MaxPooling2D(2, 2),
    keras.layers.Conv2D(64, (3, 3), activation='relu'),
    keras.layers.MaxPooling2D(2, 2),
    keras.layers.Flatten(),
    keras.layers.Dense(128, activation='relu'),
    keras.layers.Dropout(0.5),
    keras.layers.Dense(10, activation='softmax')
])
```
✓ 0.2s

```python
# Compile the model
model.compile(optimizer='adam',
              loss='sparse_categorical_crossentropy',
              metrics=['accuracy'])
```
✓ 0.0s

Fig.2 Data Preprocessing & Training

### 3. Model Evaluation:

After training, the code evaluates the model's accuracy on the test dataset. This gives an indication of how well the model performs on unseen data.

```python
# Evaluate the model on the test dataset
test_loss, test_accuracy = model.evaluate(test_images.reshape(-1, 28, 28, 1), test_labels)
```

313/313 [==============================] - 2s 8ms/step - loss: 0.0253 - accuracy: 0.9921

```python
print(f'Test accuracy: {test_accuracy * 100:.2f}%')
```

Test accuracy: 99.21%

Fig.3 Model Evaluate

### 4. Confusion Matrix:

To further assess the model's performance, the code creates a confusion matrix. This matrix helps identify which digits the model often confuses and provides insights into its strengths and weaknesses.

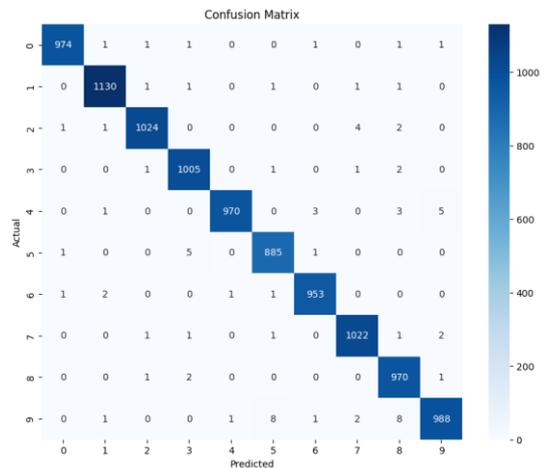

Fig.4 Confusion Matrix

**5. Real-time Prediction (Camera Input):**

The code allows real-time digit recognition using a camera feed. It captures frames from the camera using OpenCV, preprocesses each frame to match the model's input shape (28x28 pixels, normalized), and makes predictions on these frames. The predicted digit is displayed on the camera feed in real-time.

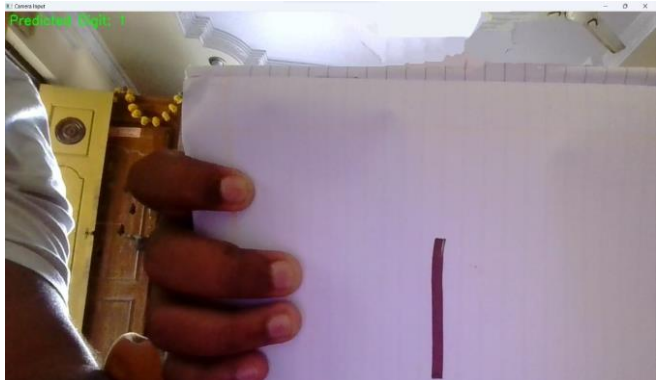

Fig.5 Real-Time Input

**6. Image Gallery Prediction:**

The code also predicts the digits from images in the device's gallery. It loads an image from the gallery, preprocesses it to match the model's input requirements, and provides the predicted digit

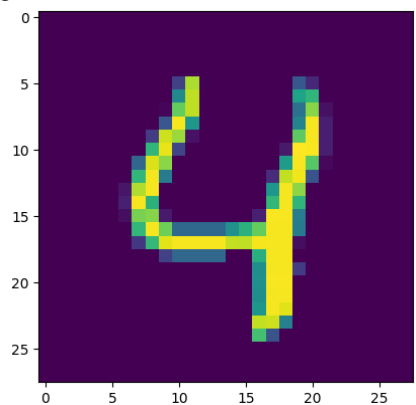

Fig.6 Predicted Digit

**7. Displaying Results:**

The model's accuracy is displayed by the code on the test dataset. The predicted digit is superimposed on the video feed for the live camera feed. The predicted digit for image gallery predictions is printed to the console.

```
image_path = 'D:\\4_1Sem\\Project\\Hand_Written_Classification\\2.jpeg'
frame = cv2.imread(image_path, cv2.IMREAD_GRAYSCALE)  # Load the image in grayscale

# Preprocess the image to match the model's input shape
frame = cv2.resize(frame, (28, 28))
frame = frame / 255.0

# Reshape the image to match the model's input shape
input_image = frame.reshape(1, 28, 28, 1)

# Use your model to make predictions
predictions = model.predict(input_image)

# Get the predicted digit (index with the highest probability)
predicted_digit = np.argmax(predictions)

print(f'Predicted Digit: {predicted_digit}')
```

```
1/1 [==============================] - 0s 56ms/step
Predicted Digit: 2
```

Fig.7 Displaying Output

## V. RESULTS AND DISCUSSIONS

MNIST handwritten digits can be recognized by our approach. Compared to the real time implementations so far, the results of our model have shown best accuracy. To predict the handwritten digits of MNIST dataset, our classification model has proven to be more precise during the prediction. For testing the model, we used a handwritten image of the MNIST dataset written by a school student while performing the prediction. Initially, the classification system will find the path of the image that has to be predicted. Then, the classification system undergoes testing and will be able to predict the image uploaded for testing with a high accuracy rate.

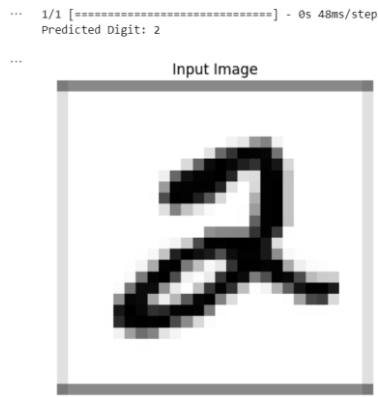

Fig.9 Result

| Reference Paper | Author | Accuracy | Precision | Recall | F-Measure |
|---|---|---|---|---|---|
| 1 | Sandeep Dwarkanath Pandea | 98.8 | 98.8 | 98.9 | 98.8 |
| 2 | Mohamed All Souibgui | 69.4 | 69.4 | 71.1 | 70.5 |
| 3 | Ali Alameer | 67.4 | 69.4 | 72.4 | 70.9 |
| 4 | M.I.I Shuvo | 58.4 | 58.4 | 63.5 | 60.8 |
| 5 | Hongge Yao | 77.2 | 77.2 | 80.1 | 78.7 |
| 6 | Fatemeh Haghighi | 98.6 | 98.6 | 98.6 | 98.6 |
| 7 | Amirreza Fatch | 73.1 | 72.5 | 75.8 | 74.1 |
| 8 | Abhinandan Chiney | 72.1 | 74.7 | 75.8 | 75.2 |
| 9 | Savita Ahlawat | 61.1 | 61.1 | 66.6 | 63.8 |
| 10 | Anisha Gupta | 80.0 | 80.0 | 82.4 | 81.1 |
| 11 | KottaKota Asish / P Sarath Teja / Kishan Chander | 97.0 | 97.6 | 97.9 | 98.0 |

Fig.10 Comparison of results

The recent studies in character recognition have shown remarkable improvements, with several researchers making significant contributions to the field. Notably, Sandeep Dwarkanath Pandea and Fatemeh Haghighi have set a new benchmark with their character recognition systems, achieving an outstanding accuracy and precision rate of 98.8%. This outstanding result reflects the continuous enhancement in the accuracy and precision of character recognition models, which is vital for applications like document analysis and automated data entry. Furthermore, KottaKota Asish's work deserves special recognition, as it boasts an impressive accuracy rate of 97.0%. These findings indicate that character recognition technology is steadily progressing and becoming increasingly reliable, holding great promise for its integration into various domains, including digitization and language processing. The collective efforts of these researchers mark a significant leap forward in the effectiveness of character recognition systems.

## VI. CONCLUSION

Our study has revealed that machine learning algorithms excel in identifying trends in diverse writing styles, particularly in recognizing handwritten digits. Neural networks stand out as highly accurate tools for this task, and their performance can be further enhanced by refining hyperparameters and eliminating ensemble features, making the model both more accurate and computationally efficient. Our research has significantly improved the accuracy of predicting student handwritten digits within the MNIST dataset, beginning with importing necessary libraries, loading the dataset, and conducting training and testing processes. The complexity of various handwriting styles presents a formidable challenge in developing a
handwritten digit identification system, with variations in angles, stress, andsegment lengths. Despite these challenges, ongoing research is focused onfine-tuning existing models and developing advanced classification algorithms to achieve accurate predictions while reducing computational costs and time investment in this field.